\documentclass[%
 reprint,
 amsmath,amssymb,
 aps,
]{revtex4-1}

\usepackage{graphicx}
\usepackage{dcolumn}
\usepackage{bm}
\usepackage{hyperref}
\usepackage{physics}

\begin{document}

\title{Is Information Theory Inherently a Theory of Causation?}

\author{David Sigtermans}
\affiliation{%
 ASML NV, The Netherlands; david.sigtermans@asml.com\\
}%

\date{\today}

\begin{abstract}
Information theory gives rise to a novel method for causal skeleton discovery by expressing associations between variables as tensors. This tensor-based approach reduces the dimensionality of the data needed to test for conditional independence, e.g., for systems comprising three variables, the causal skeleton can be determined using pair-wise determined tensors. To arrive at this result, an additional information measure, path information, is proposed.
\end{abstract}

\maketitle
The gold standard for causal inference is experimentation. Deliberately changing one variable while keeping all other variables constant, tests for three necessary conditions of a causal association: temporal precedence of the cause over the effect, the existence of a physical influence, and finally, the distinction between an apparent direct association, and a ``real'' direct association \cite{Eichler}. When experiments, or interventions, are not possible, other methods are needed to test whether the earlier mentioned conditions are met. An important class of methods are those that use graphical models reflecting the statistical (conditional) independence relations resulting from observational data \cite{Spirtes2000,Pearl}. If the data are causally sufficient, that is, there are no unobserved variables, causal associations can be inferred if the Markov property is applicable \cite{Spirtes2000}. The  vertices in the related graph represent the variables, the directed edges represent existence and directionality of the causal associations. For a graph to be causal, the faithfulness assumption \cite{Spirtes2000} must be satisfied. This assumption entails that missing edges in a probabilistic graphical model reflect the (conditional) independencies in the data. For example, in a system comprising three variables, independence of $X$ and $Z$ given $Y$, denoted as $X\!\perp \!\!\! \perp\!Z \vert Y$, implies that the causal structure is a cascade. The causal skeleton, the causal graph with undirected edges, equals $X\! \-- \! Y \! \-- \! Z$. When testing for conditional independence, a multivariate approach is therefore needed; it is deemed impossible to infer a causal skeleton using bivariate measures. An issue with multivariate approaches is the ``curse of dimensionality'' \cite{Runge}.\\ 

In this letter we summarize our work that potentially minimizes the impact of the curse of dimensionality, and enables us, in some cases, to use bivariate analysis to discover the causal skeleton for multivariate systems. Our approach is based on information theory \cite{Shannon}. An implicit assumption of this theory of communication is the existence of a mechanism that enables repeatability in communication, i.e., information theory could be used to infer causality. This idea is not new, see for example \cite{IT_CT1,IT_CT}. Mutual information, the information theoretic measure of association between random variables \cite{ThomasCover}, arises from data transmission over a noisy discrete memoryless communication channel or \emph{channel} in short. In our approach, an edge is modeled as a channel. If the maximal amount of information that channel can transfer, the so-called channel capacity \cite{ThomasCover}, equals zero, no direct causal relation can exist between the input and output of the channel, and the edge is not shown in the graph. Using an additional measure of association, path-based mutual information or \emph{path information} in short, we show that for a system comprising three variables, pair-wise determined measures can differentiate between direct and indirect associations. Because of the symmetry of mutual information, directionality cannot not follow from information theory, i.e., we can only use it to discover the causal skeleton. If one so wishes, one could use the fact that dependence between otherwise independent causes is induced when conditioning on the collider of a v-structure, e.g., $X\!\not\!\perp\!\!\!\perp\!Z \vert Y$ for the v-structure $X\! \rightarrow \! Y \! \leftarrow \! Z$. See for example the ``Fast Causal Inference'' (FCI) method described in \cite{Spirtes2000}. Here we focus on the discovery of the causal skeleton. The assumptions underlying our approach are the same assumptions used in information theory: stationarity and ergodicity of the data \cite{Shannon}. We furthermore assume that the systems under consideration are noisy.

A foundational aspect of our approach is the earlier mentioned discrete memoryless channel. This channel transforms the probability mass function    of the input data into the probability mass function of the output data via a linear transformation. In a memoryless channel, the output solely depends on its input, i.e., it encodes the Markov property. The linear map is the channel specific transition probability matrix \cite{ThomasCover}. The data are represented as discrete random variables and denoted with uppercase letters, e.g., $X$ and $Y$. The related realizations are represented by the lowercase letters, i.e., $x$ and $y$. Assuming a fixed, e.g., a lexicographical, ordering of the related sample spaces or alphabets, a one-to-one relationship exists between the realizations $x$ and $y$ and their positions in the respective alphabets. When using contra-variant and covariant notation, i.e., superscript and subscript notation, it is immediately clear from the equation whether to interpret $x$ and $y$ as indices or as realizations. With $p^x$ the probability that $X\!=\!x,\; p(x)$, $p^y$ the probability $p(y)$, and $\mathcal{A}^y_x$ the transition probability $p(y\vert x)$, the channel transforms an input probability mass function into an output probability mass function according to
 
\begin{equation} \label{eq:linearF}
	p^y = \sum_x p^x \mathcal{A}^y_x
\end{equation}

As of now, we will use the Einstein summation convention, i.e., summation is implied by indices that appear both as contra-variant and covariant indices. Eq.\,\ref{eq:linearF} can therefore be written as 

\begin{equation*}
	p^y \!=\! p^x \mathcal{A}^y_x.
\end{equation*}

When transforming the input and/or output alphabets via a linear transformation, the transition probability matrix represented using the contra-variant and covariant notation, transforms as a proper \emph{tensor} \cite{ProbTensor}. We denote a tensor with calligraphic font, e.g., $\mathcal{A}$. Let's now consider a cascade comprising three variables, the chain $X\! \rightarrow \! Y \! \rightarrow \! Z$. As we know, the mutual information for any pair of random variables, e.g., $X$ and $Z$, equals

\begin{equation} \label{eq:MI}
	I(X;Z) = \sum_{x,z} p(x,z)\log_2\left[ \frac{p(z\vert x)}{p(z)} \right].
\end{equation}

Because the mediator $Y$ is not an explicit variable in this equation, it is unknown how $Y$ influences $I(X;Z)$ in this chain. The resulting uncertainty is reflected in the data processing inequality, $I(X;Z)\!\leq\! \min [I(X;Y),I(Y;Z)]$, where $I(X;Y)$ and $I(Y;Z)$ are the mutual informations of the constituting direct associations of the chain. There is equality if one or more associations are due to a noiseless channel \cite{ThomasCover}. With observational data more information is available, allowing us to be more specific because each association is represented by a tensor, and these tensors contain the transition probabilities that can be determined from the data. We now make a small sidestep and introduce some new terminology and notation that. The tensor representing the association between the beginning and the end of this chain is referred to as ``the tensor of the path $\{X\}\{Y\}\{Z\}$''. A path with one or more mediators is called an indirect path, while a path without mediators is a so-called direct path. Using the tensors of the constituting direct paths $\{X\}\{Y\}$ and $\{Y\}\{Z\}$ denoted as $\mathcal{C}^z_x\!=\!p(z\vert x)$ and $\mathcal{B}^z_y\!=\!p(z\vert y)$ respectively, and the associated expressions for the linear transformations analogous to Eq.\,\ref{eq:linearF}, it follows that for the chain: $\mathcal{C}^z_x \!=\! \mathcal{A}^y_x \mathcal{B}^z_y$ (see Appendix \ref{App:MatrixProduct}). The mutual information resulting from transmission along the path therefore equals

\begin{equation} \label{eq:MappingPath}
	I(X;Z)_{p_{in}} = \sum_{x,z}  p^{xz} \log_2 \left[\frac{\mathcal{A}^y_x \mathcal{B}^z_y}{p^z}\right],
\end{equation}

with $p^{z}\!=\!p(z)$ and $p^{xz}\!=\!p(x,z)$. The index $p_{in}$ indicates that the probability distributions are associated to the indirect path $\{X\}\{Y\}\{Z\}$. The dynamics within the chain involving the mediator $Y$ is now captured by the matrix product $\mathcal{A}^y_x \mathcal{B}^z_y$. If the three variables in the chain $X\! \rightarrow \! Y \! \rightarrow \! Z$ are the only three variables in the system, then $I(X;Z)_{p_{in}}\!=\!I(X;Z)$. The direct association between $X$ and $Z$ is a consequence of the indirect path, so an edge between $X$ and $Z$ should be removed, in other words, the graph needs to be pruned. In Appendix \ref{App:GeneralProduct} it is proven that the existence of an additional association between $X$ and $Z$, mediated by a fourth variable, does not violate the validity of Eq.\,\ref{eq:MappingPath}, but that the path information does not equal the mutual information: $I(X;Z)_{p_{in}} \neq I(X;Z)$. Instead of comparing the information measures, we use the tensors. Inequality of the path information and the mutual information implies that $\mathcal{T}\{X\}\{Z\}$, the tensor of the direct path $\{X\}\{Z\}$, does not equal the tensor of the indirect path information: $\mathcal{T}^z_x \! \neq \! \mathcal{A}^y_x \mathcal{B}^z_y$. Therefore, comparison of the tensors of the direct and indirect paths enables, under the assumption of causal sufficiency, differentiating between direct and indirect associations using pair-wise determined tensors for a system comprising three variables. It is of course possible that the tensor of the direct association equals the tensor of the indirect association purely by chance. Using the simplified probabilistic model from Appendix \ref{App:SimpleProbabilisticModel}, it can be shown however that this probability decreases with increasing cardinalities of the alphabets because the probability scales with $10^{-N(M-1)}$. The variables $N\!\in \!\mathbb{N}$ and $M\!\in \! \mathbb{N}$ represent the cardinalities of the input and output alphabets respectively.

The approach is also applicable to forks. Consider for example the fork $Y\!\leftarrow\!X\!\rightarrow\!Z$. The indirect association between $Y$ and $Z$ is either a consequence of the path $\{Y\}\{X\}\{Z\}$, or the path $\{Z\}\{X\}\{Y\}$. Because Eq.\,\ref{eq:linearF} expresses the law of total probability \cite{LoTP}, a stochastic tensor always exists for a path, irrespective of the direction in which this path is traversed. We use the following naming convention, if the tensor of the path $\{X\}\{Y\}$ is denoted as $\mathcal{A}$, then the stochastic tensor associated to the path $\{Y\}\{X\}$ is denoted as $\mathcal{A}^{\ddagger}$, i.e., $p^x\!=\!p^y \mathcal{A}^{x\ddagger}_y$. The tensor $\mathcal{A}^{\ddagger}$ is not the inverse of $\mathcal{A}$ because the inverse of the stochastic tensor is in general not a stochastic tensor itself and therefore does not represent a channel. This naming convention allows us to write the linear transformations associated to the two paths as $p^y \mathcal{A}^{x\ddagger}_y\mathcal{C}^z_x\!=\! p^z$ and $p^z \mathcal{C}^{x\ddagger}_z\mathcal{A}^y_x\!=\!p^y$ respectively. Because a fork can be considered a chain, and the path information is independent of how the chain is traversed \cite{Sigtermans_2020}, $I(Y;Z)_{p_{in}}\!=\!I(Z;Y)_{p_{in}}$, where $p_{in}$ indicates that probability distributions are associated to the path $\{Y\}\{X\}\{Z\}$, which are identical to the probability distributions associated to the path $\{Y\}\{X\}\{Z\}$. Like mutual information, path information cannot be used to infer directionality either.
		
Extending the approach to systems comprising over three variables is straightforward because a mediator, say $\mathcal{S}$, could also be a set of variables. If the tensor of the path $\{X\}\{Z\}$ equals the tensor of the path $\{X\}\{\mathcal{S}\}\{Z\}$, the association between $X$ and $Z$ is indirect. The tensors of the paths from or towards $\mathcal{S}$ are multivariate. However, a so-called multivariate pruning step is not needed if all but one indirect paths have channel capacities that do not differ significantly from zero.\\

The concepts discussed allow us to discover the causal skeleton in three steps. First, using the observational data, an undirected graph is inferred in which each edge represents a channel capable of transferring information, i.e., only edges for which the channel capacities differ significantly from zero are shown. The channel capacity is calculated using the Blahut-Arimoto algorithm, which uses a transition probability matrix as an input \cite{Blahut}. In the next step, the resulting undirected graph is pruned using the bivariate tensors that were determined in the first step. The pruned graph is the input for the third step. In this step the channel capacity of each indirect path is calculated for any start and end point connected by a direct path and two or more indirect paths. If two or more indirect paths have a channel capacity larger than zero, the multivariate pruning step is performed. The output of the third step is the final causal skeleton.

We applied our approach to a \emph{toy} data set with a well-defined ground truth, the ``LUng CAncer Simple set'', LUCAS0 \footnote{http://www.causality.inf.ethz.ch/data/LUCAS.html}. The data in this set were generated artificially by causal Bayesian networks with binary variables and comprises 12 binary parameters, each containing 2000 samples. To infer the skeleton, the probability transition tensors and the 95\% confidence intervals were determined for all pairs of variables. For the latter, Jeffreys interval estimation for a binomial proportion was used \cite{CI}. The calculations were performed on a 2010 Mac mini with an 2.4 GHz Intel Core 2 Duo processor, and 8Gb RAM. The proposed framework was implemented in MATLAB R2018b. The processing time was 1.9 seconds.

\begin{figure} [t!]
\centering
\includegraphics[scale=0.42]{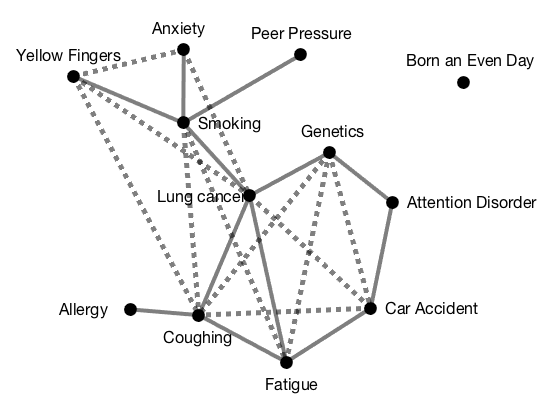}
\caption{\label{fig:LUCAS} The inferred direct and indirect associations for the LUCAS0 set after the second step. The solid lines represent the remaining direct edges. The dotted lines represent the indirect associations.}
\end{figure}

In Figure \,\ref{fig:LUCAS}, the result of the first pruning step is depicted. The direct association between \{\textsf{Lung\;Cancer}\} and \{\textsf{Fatique}\}, indicated by a solid line, is the only candidate for which a multivariate pruning step could be needed. However, the channel capacity of the indirect path of length four starting at \{\textsf{Lung\;Cancer}\} and ending at \{\textsf{Fatique}\}, and comprising direct associations only, does not differ significantly from zero: multivariate pruning is not needed. The inferred causal skeleton should therefore be equal to the skeleton of the ground truth, which is indeed the case.\\
 
To conclude, with the addition of path information, causal skeletons are a consequence of information theoretic considerations. Because directionality can be inferred from the causal skeleton using aspects from the earlier mentioned FCI method, one could argue that information theory is inherently a theory of causation. It furthermore follows from the DPI that in a noisy system the channel capacity of a cascade decreases with increasing path length, we speculate that for larger, noisy systems, the path-based method is less sensitive to the curse of dimensionality, as illustrated by our example where bivariate analysis sufficed. This will be the subject of future research. It is interesting to note that our approach is also applicable to time-series. For example, transfer entropy \cite{Schreiber} is proven to result from transmission over a set of discrete memoryless channels with associated tensors \cite{DSIG}. In this case, directionality is also inferred.\\

We acknowledge the support of Hans Onvlee, ASML Research.


\appendix
\section{Proof of Matrix Product}\label{App:MatrixProduct}
Consider the chain $X\!\rightarrow\! Y\!\rightarrow Z$, with the transition probability matrices $\mathcal{A}^y_z\!=\!p(y\vert x)$, $\mathcal{B}^z_y\!=\!p(z\vert y)$, and $\mathcal{C}^z_x\!=\!p(z\vert x)$. For this chain $\mathcal{C}^z_x \!=\! \mathcal{A}^y_x \mathcal{B}^z_y$.

Consider the linear transformation $p^z \!=\! p^y \mathcal{B}^z_y$ related to the association $Y\!\rightarrow Z$. The input for this linear transformation, $p^y$, is itself the result of the linear transformation related to the association $X\!\rightarrow\! Y$: $p^y \!=\! p^x \mathcal{A}^y_x$. Substituting the latter expression in the linear transformation  $p^z \!=\! p^y \mathcal{B}^z_y$, results in the expression

\begin{equation}
	p^z \!=\! p^x \mathcal{A}^y_x\mathcal{B}^z_y.
\end{equation}

Because the linear transformation related to the association $X\!\rightarrow\! Z$ equals $p^z \!=\! p^x \mathcal{C}^z_x$, it follows that for the chain $\mathcal{C}^z_x \!=\! \mathcal{A}^y_x \mathcal{B}^z_y$. \\

Because the transition probability matrices $\mathcal{A}^y_z\!=\!p(y\vert x)$ and $\mathcal{B}^z_y\!=\!p(z\vert y)$ are associated to the \emph{direct paths} $\{X\}\{Y\}$ and $\{Y\}\{Z\}$ respectively, the transition probability matrix $\mathcal{C}^z_x\!=\!p(z\vert x)$ is associated to the indirect path $\{X\}\{Y\}\{Z\}$. This is used in Appendix \ref{App:GeneralProduct}.

\section{Proof of Inequality Path Information and Mutual Information in Case of Multiple Indirect Paths} \label{App:GeneralProduct}
Consider system comprising the chains $X\!\rightarrow\! Y\!\rightarrow Z$ and $X\!\rightarrow\! U\!\rightarrow Z$. Because the product $\mathcal{A}^y_x \mathcal{B}^z_y\!=\! \mathcal{C}^z_x$ is associated to path $\{X\}\{Y\}\{Z\}$,
	
\begin{equation*} 
	I(X;Z)_{p_{in}} = \sum_{x,z}  p^{xz} \log_2 \left[\frac{\mathcal{A}^y_x \mathcal{B}^z_y}{p^z}\right],
\end{equation*}

where index $p_{in}$ indicates that the probability distributions are associated to the indirect path $\{X\}\{Y\}\{Z\}$. Because $X \not\!\perp \!\!\! \perp Z \vert Y$, the product $\mathcal{A}^y_x \mathcal{B}^z_y$ does not equal $p(z\vert x)$, i.e., 

\begin{equation*} 
	I(X;Z)_{p_{in}} \neq I(X;Z),
\end{equation*}

with $p_{in}$ indicating that the probability distributions for the left-hand side of the equation are associated to the indirect path $\{X\}\{Y\}\{Z\}$.

\section{Simple Probabilistic Model for Coincidental Equality of Direct and Indirect Associations}\label{App:SimpleProbabilisticModel}
Let's assume that the tensor for the indirect path is a stochastic $N\!\times\! M$ tensor, where $N\!\in \!\mathbb{N}$ and $M\!\in \! \mathbb{N}$ represent the cardinalities of the input and output alphabets, respectively. Per element of this tensor, an associated sample space is determined, consisting of integers $\geq 1$. The cardinality of the sample space is equal to the product of each tensor element and $10^{\varsigma}$, where $\varsigma\!\in\!\mathbb{N}$ represents the number of significant digits. Finally, a new $N\!\times\! M$ transition probability tensor is constructed by selecting per tensor element a number at random from the associated sample space, and multiplying it by $10^{-\varsigma}$. The probability that this newly constructed stochastic tensor equals the tensor of the indirect association $p(\varsigma,N,M)$ scales with $10^{-\varsigma \cdot N(M-1)}$. For binary data, this probability is small for $\varsigma \!\geq\! 2$, e.g., $p(2,2,2)\! \approx \! 10^{-4}$. The probability approaches zero rapidly with increasing cardinalities.  The probability that the tensor of the direct association is equal to the tensor of the indirect association purely by chance is therefore negligible.


\end{document}